\def\year{2020}\relax
\documentclass[letterpaper]{article} 
\usepackage{aaai20}  
\usepackage{times}  
\usepackage{helvet} 
\usepackage{courier}  
\usepackage[hyphens]{url}  
\usepackage{graphicx} 
\usepackage{graphicx}  

\usepackage{amssymb}
\usepackage{amsmath}
\usepackage{colortbl}
\usepackage{multirow}
\usepackage{subfigure}
\usepackage{xcolor}
\usepackage{algorithm}
\usepackage{algorithmic}

\DeclareMathOperator*{\argmin}{argmin}

\title{Constructing Multiple Tasks for Augmentation: \\Improving Neural Image Classification With $K$-means Features}
\author{\Large \textbf{Tao Gui\thanks{Both authors contributed equally}, Lizhi Qing$^*$, Qi Zhang, Jiacheng Ye, HangYan, Zichu Fei, Xuanjing Huang} \\ Shanghai Key Laboratory of Intelligent Information Processing, Fudan University \\School of Computer Science, Fudan University \\ 825 Zhangheng Road, Shanghai, China \\ \{tgui16, lzqing18, qz, yejc19, hyan19, zcfei19, xjhuang\}@fudan.edu.cn}
\begin{document}

\maketitle

\begin{abstract}


Multi-task learning (MTL) has received considerable attention, and numerous deep learning applications benefit from MTL with multiple objectives. However, constructing multiple related tasks is difficult, and sometimes only a single task is available for training in a dataset. To tackle this problem, we explored the idea of using unsupervised clustering to construct a variety of auxiliary tasks from unlabeled data or existing labeled data. We found that some of these newly constructed tasks could exhibit semantic meanings corresponding to certain human-specific attributes, but some were non-ideal. In order to effectively reduce the impact of non-ideal auxiliary tasks on the main task, we further proposed a novel meta-learning-based multi-task learning approach, which trained the shared hidden layers on auxiliary tasks, while the meta-optimization objective was to minimize the loss on the main task, ensuring that the optimizing direction led to an improvement on the main task. Experimental results across five image datasets demonstrated that the proposed method significantly outperformed existing single task learning, semi-supervised learning, and some data augmentation methods, including an improvement of more than 9\% on the Omniglot dataset. 
\end{abstract}

\section{Introduction}
Multi-task learning \cite{caruana1997multitask} is a learning paradigm in machine learning. Its goal is to improve the generalization performance of a single task by leveraging useful information contained in multiple related tasks \cite{zhang2017survey}. Multi-task learning has been used successfully across all applications of machine learning, from natural language processing \cite{collobert2008unified} and speech recognition \cite{deng2013new} to computer vision \cite{girshick2015fast} and reinforcement learning \cite{wilson2007multi}. Critical assumptions in these applications are that MTL typically involves very heterogeneous tasks and the tasks are closely related in some way \cite{ruder2017overview,zhang2017survey}.


\begin{figure}[t]
\centering
  \includegraphics[width=2.4in]{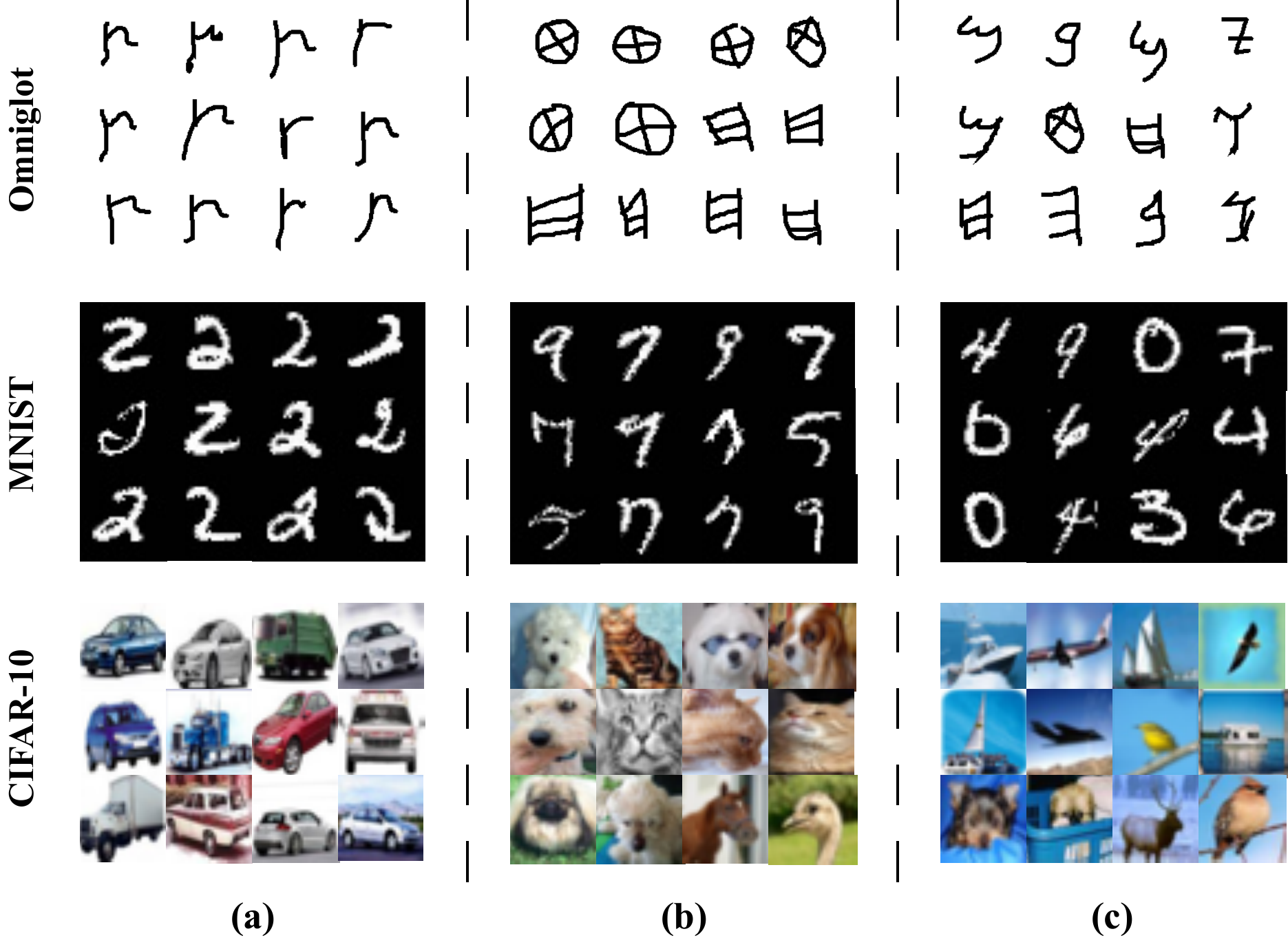}
  \caption{Examples of $k$-means clustering classes based on ACAI embeddings with random scaling \cite{berthelot2018understanding}, which can be used to perform multi-task learning. (a) Some of the clusters correspond well to true labels. (b) Some of the clusters exhibit semantic meaning, like the heads of the animals. (c) Others are non-ideal or are based on image artifacts, which could cause a performance deterioration. We proposed a meta-learning-based MTL to tackle this problem.}
  \label{fig:example}
\end{figure}

Although the use of MTL to improve single task learning (STL) has achieved great success \cite{luong2015multi,yim2015rotating}, these methods continue to have difficulty with two issues. \textbf{First}, we have always been hindered in searching a sufficient number of related tasks \cite{bingel2017identifying}. In particular, in some domains, only a single task is available for training. \textbf{Second}, because of the lack of a good notion of when two tasks should be considered related \cite{ruder2017overview}, it is difficult to avoid introducing some unrelated tasks, which can result in the learned features being negatively influenced \cite{zhang2017survey}. \citeauthor{meyerson2018pseudo} \shortcite{meyerson2018pseudo} explored the pseudo-task augmentation method, which verified that MTL improved the generalization performance of single task. However, they only regarded the duplicates of a single task as multiple tasks, which lost the diversity of the task and could not take advantage of unlabeled data.



In this work, we considered an extreme situation where only a single task was available for training, but we expected to find sufficient related tasks and utilize the framework of deep MTL to improve the single task learning. To tackle the above issues, we used unsupervised clustering methods on the unlabeled data or existing labeled data to automatically construct auxiliary tasks. Because the unsupervised clustering methods are fast implemented on a large scale, we can easily to obtain sufficient tasks. As for the \textit{relatedness} of tasks, because we constructed the auxiliary tasks from the same data distribution as the main task, all of the tasks were $\mathcal{F}$-related and could yield provable multiple-task learning guarantees \cite{ben2008notion}\footnote{\citeauthor{ben2008notion} \shortcite{ben2008notion} proposed that two tasks were $\mathcal{F}$-related if the data for both tasks could be generated from a fixed probability distribution using a set of transformations $\mathcal{F}$. We constructed the tasks based on the same data distribution, which was a special case of $\mathcal{F}$-related.}. We found that with simple $k$-means clustering mechanisms for partitioning the embedding space \cite{bojanowski2017unsupervised,donahue2016adversarial}, some of the clusters could construct reasonable related auxiliary tasks \cite{coates2012learning}.  As shown in Figure \ref{fig:example} (a) and (b), some of the clusters correspond well to unseen labels or exhibit semantic meaning. Hence, these related tasks could benefit the MTL \cite{bonilla2008multi}.

Unsupervised clustering would inevitably produce the non-ideal clusters, as shown in Figure \ref{fig:example} (c), which could cause a performance deterioration. To tackle this problem, we introduced a meta-learning-based multi-task learning framework (Meta-MTL), an novel variant of Model-Agnostic Meta-Learning (MAML) \cite{finn2017model}, to assist in the fast and robust adaptation of the MTL model, which drove the training of any auxiliary tasks to be beneficial to the main task. Specifically, Meta-MTL adopted a hard parameter sharing framework, which trained the shared hidden layers on a set of auxiliary tasks, while the meta-optimization objective was to minimize the loss on the main tasks, ensuring that the optimizing direction led to an improvement on the main task.

The main contributions of this paper can be summarized as follows: 1) we studied the multi-task learning in an extreme situation where only a single task was available, and used the $k$-means clustering to conduct multi-task learning to improve STL; 2) we proposed a novel meta-learning-based MTL method to assist in the fast and robust adaptation for the main task, which could effectively avoid the influence of the non-ideal clusters; and 3) experimental results across five image datasets indicated that the proposed method significantly outperformed the STL, semi-supervised learning, and some data augmentation methods.

\section{Related Work and Background}
This work is related to two research threads: multi-task learning and meta-learning. We first summarize the related work and then give some background. We also discuss the differences between the existing approaches and our own.

\subsection{Multi-Task Learning}
Many deep learning approaches require a large number of training samples and are becoming increasingly complex \cite{krizhevsky2012imagenet,simonyan2014very,he2016deep}. However, this requirement cannot be fulfilled in some applications because (labeled) samples are hard to collect. MTL is a good solution to this insufficient data problem when there are multiple related tasks each of which has limited training samples \cite{zhang2017survey}.

In MTL, a task $\mathcal{T}_t$ is usually accompanied by a training dataset $D_t$ consisting of $N_t$ training samples, i.e., $D_t = \{\mathbf{x}_i^t, \mathbf{y}_i^t\}_{i=1}^{N_t}$, where $\mathbf{x}_i^t \in \mathbb{R}^{d_t}$ is the $i$th training instance in $\mathcal{T}_t$ and $y_i^t$ is its label. Although more sophisticated methods now exist, the most common approach is still based on the joint training of neural network models for multiple tasks \cite{caruana1997multitask}, in which a joint model is decomposed into a feature extractor (shared layers) $\mathcal{F}$ that is shared across all tasks, and task-specific decoder $\mathcal{D}_t$ for each task. The model for the $t$th task is then defined as
\begin{equation}
\hat{\mathbf{y}}_i^t = \mathcal{D}_t(\mathcal{F}(\mathbf{x}_i^t;\theta_\mathcal{F});\theta_{\mathcal{D}_t}),\label{base}
\end{equation}
where $\theta_\mathcal{F}$ and $\theta_{\mathcal{D}_t}$ are the parameters of $\mathcal{F}$ and $\mathcal{D}_t$, respectively. Then, the goal of the joint model is to find optimal parameters $\theta^* = (\{\theta_{\mathcal{D}_t}\}_{t=1}^T, \theta_\mathcal{F})$ such that
\begin{equation}
\theta^* = \argmin_\theta \frac{1}{T}\sum_{t=1}^T\frac{1}{N_t}\sum_{i=1}^{N_t}\mathcal{L}(\hat{\mathbf{y}}_i^t, \mathbf{y}_i^t),
\end{equation} 
where $\mathcal{L}$ is a suitable loss function. More sophisticated deep MTL approaches can be characterized by ``how to share,'' which specifies concrete ways to share knowledge among tasks, such as the low-rank approach \cite{han2016multi}, task relation learning \cite{long2017learning}, and so on. In this work, we consider a method to take advantage of the benefits of MTL to improve the STL, where all of the auxiliary tasks are generated by unsupervised clustering methods. Ideally, another potential benefit is that all of the MTL techniques can be used on our models.

\begin{figure*}[t]
\centering
  \includegraphics[width=5.5in]{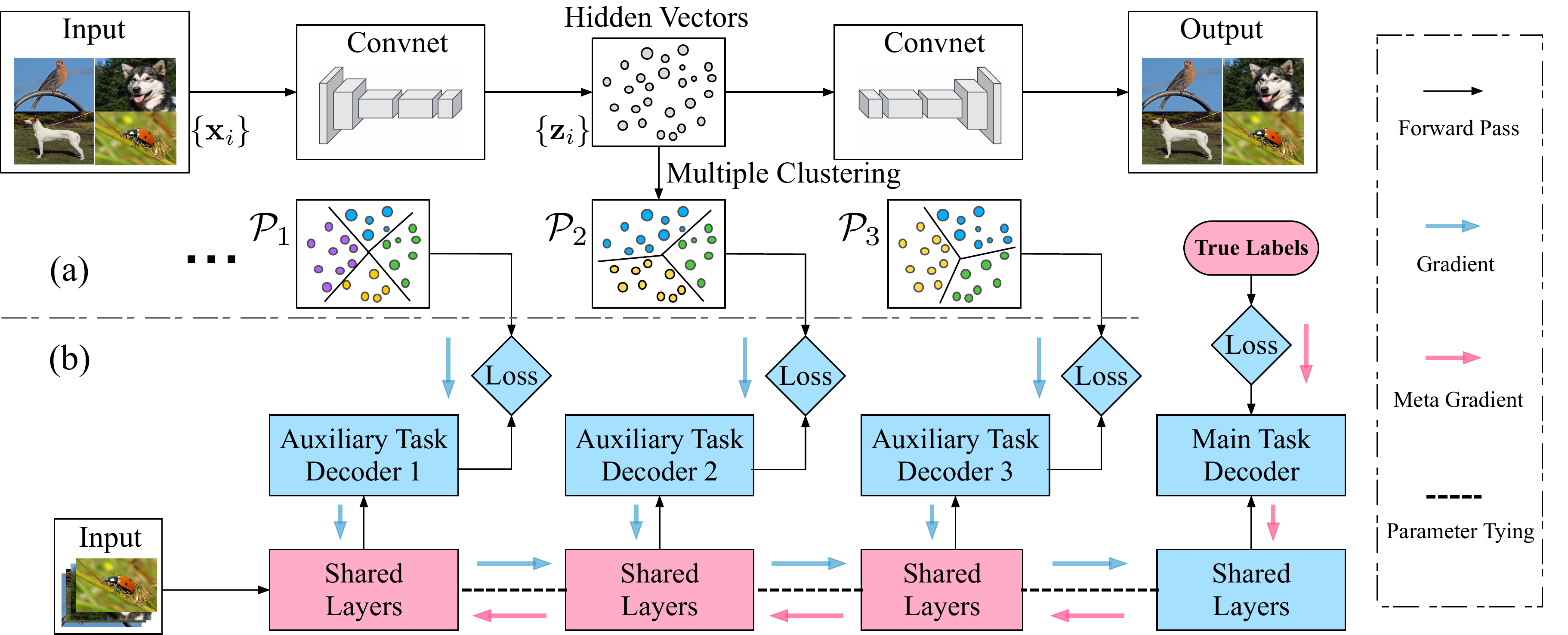}
  \caption{Graphical illustration of training process for proposed Meta-MTL with k-means augmentation. The entire training process is divided into two steps. (a) In the task generation step, we first run an unsupervised embedding learning algorithm to map the data into the embedding space $\mathcal{Z}$, producing $\{\mathbf{z}_i\}$. Then, we apply $k$-means $T$ times to obtain $T$ auxiliary tasks. (b) In the meta multi-task learning step, for each episode, one batch of images are sampled to update the task-specific decoders, while the meta-optimization objective of the shared layers is to minimize the loss on the main task. The blue boxes and arrows are mainly involved in task-specific learning, and those in purple are mainly involved meta-learning.}
  \label{fig:model}
\end{figure*}

\subsection{Meta-Learning}
Meta-learning was recently proposed and shown to achieve great performances on various few-shot learning tasks \cite{lake2015human}. The intuition of meta-learning can be categorized by learning a matrix function that embeds data in the same class closer to each other \cite{vinyals2016matching,snell2017prototypical}, learning good weight initializations \cite{finn2017model}, and learning transferable optimizers \cite{andrychowicz2016learning}. In this work, we adopt the idea from meta-learning for task generalization \cite{li2018learning}, which is a novel variant of MAML for quickly and robustly adapting the newly constructed tasks to the main task. The proposed method utilizes a new meta-learning objective on shared layers that simulates transfer learning by fine-tuning, i.e., the shared layers are updated to minimize the loss on the main task based on a few examples and a few gradient descent steps on auxiliary tasks, while the task-specific decoders are updated to optimize the corresponding tasks. Our method can be applied to most deep STL architectures.

\section{Meta-MTL with $K$-means Augmentation}
In this work, we propose a novel MTL framework with a $k$-means augmentation framework to automatically construct auxiliary tasks to improve single task learning, which is a modification to the basic MTL architecture. By using a novel meta-learning method for task generalization, the proposed method can quickly and robustly adapt the newly constructed tasks to the main task. The entire dataset construction and training process is illustrated in Figure \ref{fig:model}.

\subsection{Problem Statement}
Assume that the human-specific main task to be learned is a classic supervised classification task. We have access to a small labeled dataset $D_{main}$ and an unlabeled dataset $D_{aux}$. We expect to use the unsupervised clustering method on $D_{aux}$ to effectively construct multiple auxiliary tasks $\{\mathcal{T}_t\}_{t=1}^T$. If we lack $D_{aux}$, it is also feasible to use $D_{main}$ to construct auxiliary tasks. Our goal is to leverage the MTL framework on the automatically constructed tasks to improve the single task learning.

Concretely, we adopt a hard parameter sharing framework, which shares the hidden layers between the main task and all auxiliary tasks to obtain the shared features $\mathbf{h} = \mathcal{F}(\mathbf{x})$, while keeping several task-specific decoders to obtain task-specific outputs $\hat{\mathbf{y}}_t = \mathcal{D}_t(\mathbf{h})$. Given a main task and $T$ newly constructed auxiliary tasks, the goal of joint training all of the tasks is to minimize the total loss, similar to that of the traditional MTL, as follows:
\begin{equation}
\theta^* = \argmin_\theta\frac{1}{N_0}\sum_{i=1}^{N_0}\mathcal{L}(\hat{\mathbf{y}}_i^0,\mathbf{y}_i^0)+ \sum_{t=1}^T\frac{1}{N_t}\sum_{i=1}^{N_t}\mathcal{L}(\hat{\mathbf{y}}_i^t, \mathbf{y}_i^t)),
\end{equation}
where $\theta = (\{\theta_{\mathcal{D}_t}\}_{t=0}^T,\theta_\mathcal{F})$. $\mathcal{D}_0$ refers to the output decoder of the main task, and $\{\mathcal{D}_t\}_{t=1}^T$ refers to the decoders of $T$ auxiliary tasks. \citeauthor{baxter1997bayesian} \shortcite{baxter1997bayesian} showed that the risk of overfitting the shared parameters is an order $T$ smaller than that of overfitting the task-specific parameters. Thus, when learning more tasks simultaneously, there is less chance of overfitting on the main task, which can be verified in Section \ref{cifar}. In the next section, we discuss how we can construct tasks without ground-truth labels. In addition, we introduce a meta-learning-based MTL to assist in the fast and robust adaptation for the main task in Section \ref{MAML}.

\subsection{Automatic Task Generation}
\label{kmeans}
We aim to construct multiple classification tasks from the unlabeled data and then learn how to efficiently utilize MTL to improve STL. If such tasks are adequately structured and related, then joint learning these tasks should make it possible to improve the learning of human-provided main tasks.

Notice that in the labeled dataset, the labels $\mathbf{y}$ can induce a partition $\mathcal{P} = \{C_l\}_{l=1}^L$ over $D_0$ by assigning all of the datapoints with label $l$ to subset $C_l$. Then, subsets $\{C_l\}_{l=1}^L$ make up the entire data set, where $L$ is the number of types of labels. Similarly, we can reduce the problem of constructing multiple auxiliary tasks to that of constructing different partitions over $D_{aux}$. However, simple partitioning methods may fail. For example, randomly partitioning the data shows no consistency between a task's training data and test data. Meanwhile, clustering in the pixel-space is also unappealing because of the poor correlation between the distances in the pixel-space and the clustering difficulty caused by the high dimensionality of the raw images \cite{hsu2018unsupervised}. Hence, all that’s left is to find a reasonable alternative to human labels for defining the partition. 

Fortunately, for certain architectures, like autoencoder, the latent embeddings have been shown to disentangle important factors of variation in the dataset, which makes such models useful for representation learning \cite{berthelot2018understanding} and have been used for producing pseudo tasks with semantic meaning \cite{hsu2018unsupervised}. Thus, we first run an unsupervised embedding learning algorithm $\mathcal{E}$ on $D_{aux}$, and then map the data $\{\mathbf{x}_i\}$ into the embedding space $\mathcal{Z}$, producing $\{\mathbf{z}_i\}$. We focus on a standard clustering algorithm, $k$-means, which takes a set of vectors as input, and clusters them into $k$ distinct groups based on a geometric criterion. To produce a diverse task set, we apply random scaling to the dimensions of $\mathcal{Z}$ or random select half of the dimensions $T$ times to induce $T$ different matrices, and then generate $T$ partitions $\{\mathcal{P}_t\}_{t=1}^T$ by running $k$-means $T$ times. More precisely, for a single run of clustering, it jointly learns a $d \times k$ centroid diagonal matrix $\mathbf{U}$ and the cluster assignments $\mathbf{y}_l^*$ of each vector $\mathbf{z}_i$ such that
\begin{equation}
\mathcal{P}, \mathbf{U}  = \argmin_{\mathbf{U}\in \mathbb{R}^{d\times k}}\frac{1}{N}\sum_{i=1}^N\argmin_{\mathbf{y}_l^*\in\{0,1\} ^k}\parallel\mathbf{z}_i - \mathbf{U}\mathbf{y}_l^*\parallel^2,
\end{equation}
where $\mathbf{y}_l^{*\top}1_k=1$. Through the above function, we can obtain a set of optimal assignments $\{\mathbf{y}_l^*\}$ and a centroid matrix $\mathbf{U}$. These assignments are then used as pseudo-labels to construct the auxiliary task. As a result, the $T$ partitions can correspond to $T$ auxiliary tasks.

\subsection{Meta-MTL with Newly Constructed Tasks}\label{MAML} 
So far, we have obtained one main task and $T$ auxiliary tasks. Next, we can directly apply the MTL framework to learn a joint neural classification function $\hat{\mathbf{y}}_i^t = \mathcal{D}_t(\mathcal{F}(\mathbf{x}_i;\theta_\mathcal{F}), \theta_{\mathcal{D}_t})$. However, this can be problematic when there are some non-ideal clusters, as shown in Figure \ref{fig:example} (c), which could cause a performance deterioration \cite{zhang2017survey}. For example, ACAI’s pixel-wise reconstruction loss may prioritize information about the few ``features'' that dominate the reconstruction pixel count in complex images, resulting in clusters that only correspond to a limited range of factors such as the background color and pose \cite{hsu2018unsupervised}.

\begin{algorithm}[t]
\begin{algorithmic}[1]
\STATE Run embedding learning algorithm $\mathcal{E}$ on $D_{aux}$ and produce embeddings $\{\mathbf{z}_i\}$ from observations $\{\mathbf{x}_i\}$.
\STATE Run $k$-means on $\{\mathbf{z}_i\}$ $T$ times (with random scaling or random selection on dimensions) to generate a set of partitions $\{\mathcal{P}_t = \{C^l\}_{l=1}^{L_t}\}_{t=1}^T$, which correspond to a set of auxiliary tasks $\{\mathcal{T}_t\}_{t=1}^T$.
\FOR{episode = 1, M}
\STATE Sample batch of tasks $\mathcal{T} \sim \{\mathcal{T}_t\}_{t=0}^T$.
  \FOR{\textbf{all} $\mathcal{T}$}
  \STATE Sample $K$ datapoints $D_\mathcal{T} = \{\mathbf{x}_j,\mathbf{y}_j\}$.
  \STATE Evaluate $\nabla_{\theta_\mathcal{F}}$ and $\nabla_{\theta_{\mathcal{D}_t}}$ using $D_\mathcal{T}$ based on Equation \ref{base}.
  \STATE Applying gradient decent to update the parameters of task-specific decoders $\theta_{\mathcal{D}_\mathcal{T}}$.
  \STATE Compute updated parameters $\theta_{\mathcal{F}}^*$ with gradient descent based on Equation \ref{meta-train}.
  \STATE Sample datapoints $D_0 = \{\mathbf{x}_j,\mathbf{y}_j\}$ from $\mathcal{T}_0$ for the meta-update.
  \ENDFOR
  \STATE Update the parameters of shared layers $\theta_{\mathcal{F}}$ based on Equation \ref{meta-test}.
\ENDFOR
\end{algorithmic}
\caption{\textbf{Meta-MTL with $K$-means Augmentation}}
\label{alg:Meta-MTL}
\end{algorithm}

Therefore, to further improve the performance, we adopt a novel meta-learning-based multi-task learning method to achieve fast and robust task generalization. Instead of simply joint training $\mathcal{D}_t(\mathcal{F}(\mathbf{x}_i;\theta_\mathcal{F}), \theta_{\mathcal{D}_t})$, our method provides a way of leveraging the auxiliary tasks for better generalization on the main task in the MTL phase. More specifically, our method adopts a traditional hard sharing parameter structure, which trains each task-specific decoder $\mathcal{D}_t$ for its respective task while training the shared layers $\mathcal{F}$ on the auxiliary tasks to generalize to main task. In each training episode, we first uniformly sample a batch of tasks $\mathcal{T}$ from both the auxiliary tasks and main task $\{\mathcal{T}_t\}_{t=0}^T$, and conduct gradient descent based on the data from the sampled tasks $\mathcal{T}$ to optimize the task-specific decoders $\theta_{\mathcal{D}_\mathcal{T}}$. We simultaneously store the updated weights of shared layers $\theta_{\mathcal{F}}^*$ in preparation for the follow-up meta learning. For simplification, we use $\mathcal{L}$ to denote the loss function of our objective function. The update process is as follows:
\begin{equation}
\begin{aligned}
\theta_{\mathcal{D}_\mathcal{T}} &= \theta_{\mathcal{D}_\mathcal{T}} - \alpha\nabla\mathcal{L}_{\mathcal{D}_\mathcal{T}}(\theta_{\mathcal{D}_\mathcal{T}})  \\
\theta_\mathcal{F}^* &= \theta_\mathcal{F} - \alpha\nabla\mathcal{L}_{\mathcal{D}_\mathcal{T}}(\theta_\mathcal{F}).
\end{aligned}\label{meta-train}
\end{equation}
We then treat $\theta_\mathcal{F}^*$ as an initialized shared weight to optimize the original shared parameters on the main task based on the newly sampled datapoints in $D_0$. The final update on the shared parameters in each training episode can be written as follows:
\begin{equation}
\begin{aligned}
\theta_\mathcal{F} &= \theta_\mathcal{F} - \beta\nabla\theta_\mathcal{F}\mathcal{L}_{\mathcal{D}_0}(\theta_\mathcal{F}^*)\\
&= \theta_\mathcal{F} - \beta\nabla\theta_\mathcal{F}\mathcal{L}_{\mathcal{D}_0}(\theta_\mathcal{F} - \alpha\nabla\mathcal{L}_{\mathcal{D}_\mathcal{T}}(\theta_\mathcal{F})),
\end{aligned}\label{meta-test}
\end{equation}
where both $\alpha$ and $\beta$ are hyper-parameters of the two-stage learning rate. The above optimization can be conducted with stochastic gradient descent (SGD). We detail the task construction and Meta-MTL algorithm in Algorithm \ref{alg:Meta-MTL}. In the meta-train stage, we optimize the task-specific decoders based on the sampled tasks $\mathcal{T}$, which ensures that the model has the ability for each task. Different from decoders, in the meta-test stage, we only optimize the shared layers based on the data from the main task, which ensures that the MTL optimizing direction leads to an improvement on the main task. In this way, the knowledge learned from $\mathcal{D}_t$ can provide a good initial representation that can be effectively fine-tuned using a few examples in $\mathcal{D}_0$, and thus achieves fast and robust adaptation for the main task.

\section{Experiments}
In this section, we aim to compare the proposed method with various baseline methods on five benchmarks. To avoid falsely embellishing the capabilities of our approach by overfitting to the specific datasets and task types, we did not perform any architecture engineering: we used architectures from prior work as-is, or lightly adapted them to our needs if necessary. We designed the experiments and showed the superiority in a range of settings: (1) the Meta-MTL can achieve a significant improvement in a limited training data setting; (2) the performance of the proposed model can be further improved when combined with unlabeled data; (3) the proposed method can improve the performance of the model with data augmentation, and outperforms some semi-supervised methods; (4) the proposed method can also be used in challenging recent computer vision benchmarks, such as CIFIA-100 and miniImageNet. For the sake of simplicity, we used the same number of $k$-means clusters as that of the types of true labels for each dataset. Our code will be released at \textit{https://github.com/Howardqlz/Meta-MTL.}

\begin{table}[t]
\centering
\scalebox{0.91}{
\begin{tabular}{lc}
  \hline
  \textbf{Method}  &  \textbf{Acc.}\\
  \hline
  STL \cite{yang2016deep} & $65.72$ \\
  ACAI embedding finetune \cite{berthelot2018understanding} & $67.91$ \\
  MTL on all alphabets \cite{yang2016deep} & $70.98$ \\
  MTL + Tasks with random labels, $T = 4$ & $60.98$ \\
  MTL + Tasks with $k$-means labels, $T = 4$ & $61.26$ \\
  PTA-F, $T = 4$ \cite{meyerson2018pseudo} & $70.63$ \\
  PTA-F, $T = 10$ \cite{meyerson2018pseudo} & $71.52$ \\
  \hline
  \textbf{Meta-MTL}, $T = 4$  & $\mathbf{72.04}$ \\
  \textbf{Meta-MTL}, $T = 10$  & $\mathbf{74.80}$ \\
  \hline
\end{tabular}}
\caption{Accuracy on Omniglot dataset. THe test accuracy averaged across 50 alphabets is shown.}
  \label{tab:Omniglot}
\end{table}

\subsection{Omniglot Character Recognition}
The Omniglot dataset \cite{lake2015human} contains handwritten characters in 50 different alphabets. Each alphabet with its own number of unique characters ($14\sim55$) induces its own character recognition task. In total, there are 1,623 unique characters, and each has exactly 20 instances. Here, each task corresponds to an alphabet, and the goal is to recognize its characters. To reduce variance and improve the reproducibility of the experiments, we used the same 50/20/30\% train/validation/test split as \citeauthor{meyerson2018pseudo} \shortcite{meyerson2018pseudo} for each task. Methods were evaluated with respect to all 50 tasks. The underlying model $\mathcal{F}$ for all the baselines is a simple four layer convolutional network that has been shown to yield good performance on Omniglot \cite{yang2016deep}. Each of these four convolutional layers has 53 filters and 3$\times$3 kernels, and is followed by a 2$\times$2 maxpooling layer and dropout layer with a 0.5 dropout probability. Each task-specific decoder $\mathcal{D}_t$ has two fully connected layers with 848 and a specific number of classes of neurons, respectively. In our method, we used the $k$-means method on ACAI embeddings \cite{berthelot2018understanding} to obtain the auxiliary tasks.

Table \ref{tab:Omniglot} reports the average accuracy values across all 50 tasks (alphabets). Our proposed Meta-MTL methods surpass the STL model by more than 9\%. Not all of the baseline models work well when the training dataset is very small. \citeauthor{meyerson2018pseudo} \shortcite{meyerson2018pseudo} constructed an MTL method using duplicates of a single task, but did not introduce related tasks, which limited the performance\footnote{In Table 1$\sim$5, we evaluated six kinds of explicit controls to PTA trajectories \cite{meyerson2018pseudo}. The PTA-F reported in these tables is the best one.}. In particular, constructing tasks with random labels also resulted in a huge performance degradation. In contrast, our method can achieve a significant improvement when the main task leverages auxiliary tasks constructed by unsupervised clustering on ACAI embeddings.

\begin{table}[t]
\centering
\scalebox{0.82}{
\begin{tabular}{lc}
  \hline
  \textbf{Method}  &  \textbf{Acc.}\\
  \hline
  STL \cite{yang2016deep} & $91.83$ \\
  ACAI embedding finetune \cite{berthelot2018understanding} & 93.09 \\
  Self-training \cite{rosenberg2005semi} & 92.20 \\
  Co-training \cite{chen2011co} & 91.87  \\
  MTL + Tasks with random labels, $T = 4$ & $92.27$ \\
  MTL + Tasks with $k$-means labels, $T = 4$ & $92.86$ \\
  PTA-F, $T = 4$ \cite{meyerson2018pseudo} & $92.67$ \\
  PTA-F, $T = 10$ \cite{meyerson2018pseudo} & $91.98$ \\
  \hline
  \textbf{Meta-MTL}, $T = 4$ & $\mathbf{93.37}$ \\
  \textbf{Meta-MTL}, $T = 10$ & $\mathbf{94.22}$ \\
  \hline
  \textbf{Meta-MTL}, $T = 4$$\dagger$ & $\mathbf{93.76}$ \\
  \textbf{Meta-MTL}, $T = 10$$\dagger$ & $\mathbf{94.42}$ \\
  \hline
\end{tabular}}
\caption{Accuracy on MNIST dataset. All of the models use 1\% training data. The models marked with $\dagger$ use the remaining unlabeled data.}
  \label{tab:MNIST}
\end{table}

\begin{figure*}[t]
\centering
  \includegraphics[width=6in]{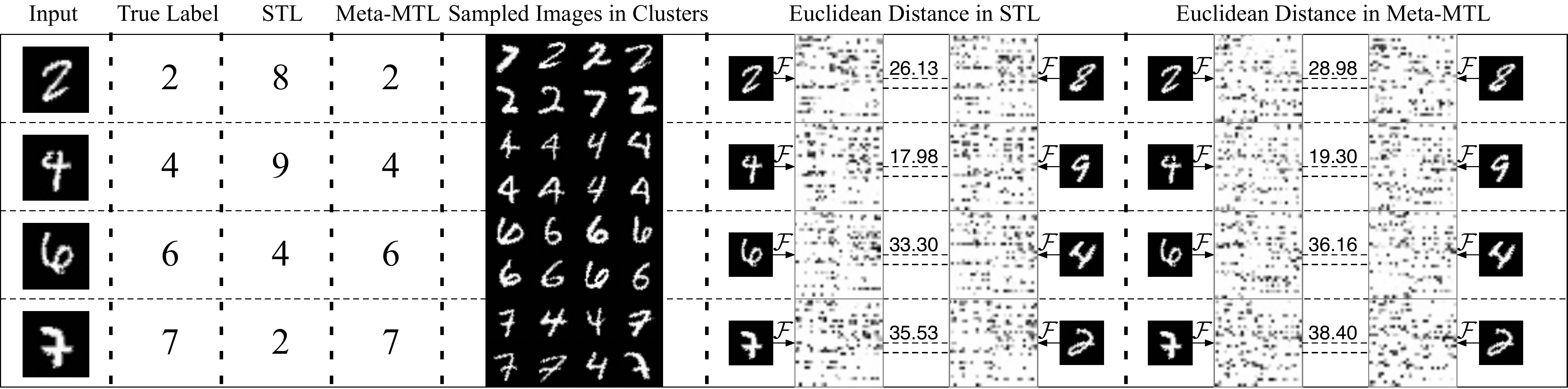}
  \caption{Visualization of behaviors of STL and Meta-MTL. The third and fourth columns are the labels predicted by STL and Meta-MTL, respectively. We randomly sample 8 images from the $k$-means cluster which contains the original image shown in the fifth column. In addition to exhibiting the output matrices of the shared layers, we also compute the Euclidean distance between the output vectors. The larger Euclidean distance shows stronger capacity to distinguish the ambiguous images.}
  \label{fig:visualization}
\end{figure*}

\subsection{MNIST Number Classification}
The MNIST dataset consists of 70,000 hand-drawn examples of the 10 numerical digits. The previous split used the original MNIST 60,000/10,000 training/testing split. In this work, we evaluated the proposed method using only 1\% training data, and the remaining data were treated as unlabeled data. Because it is a simpler task, we used two layers of the CNN architecture as the shared layer $\mathcal{F}$. The first convolutional layer has 32 filters of size 5$\times$5, followed by 2$\times$2 max pooling. The second convolutional layer has 64 filters of size 4$\times$4, and again a 2$\times$2 max pooling, as shown in \cite{yang2016deep}.

Table \ref{tab:MNIST} shows the accuracy of the baseline models and our model on the test dataset of MNIST. Similar to Omniglot, our method outperforms all of the baseline models when the training dataset is very small. In particular, when the number of the auxiliary tasks increases, the performance of PTA-F drops, but our model performance improves a lot. In addition, if we used the remaining unlabeled data to train the embeddings, our method can obtain better results. The Meta-MTL model also outperforms semi-supervised methods, such as the self-training and co-training methods by more than 3\%.

To investigate why the proposed method can improve the STL, we visualized the behaviors of the STL and Meta-MTL when encountering somewhat ambiguous hand-drawn examples. We trained Meta-MTL using the same hyperparameters as the STL, while adding only one auxiliary task to improve the STL. Some typical examples are shown in Figure \ref{fig:visualization}. From the figure, we can see that with a small training dataset, the STL is more likely to confuse some ambiguous samples, while the auxiliary task can help the model recognize them. For instance, as shown in the first row, the STL is confused if this picture is ``2'' or ``8,'' while the auxiliary cluster can give key information that this image is gathered with many examples of ``2.'' Then, the Meta-MTL can correct the STL's mistakes. In addition, we also visualized the outputs of the shared layers when encountering some ambiguous images. We resized the 1,024 dimensional output vectors to the matrices of size 32$\times$32, and then plotted the largest 200 values in the matrices. Comparing these pictures, we can observe that the outputs of the ambiguous images in STL have greater similarity than those in Meta-MTL. Therefore, the STL is more difficult to distinguish the ambiguous images. For a clearer comparison, we computed the Euclidean distance between the ambiguous images. We can see that our Meta-MTL can more easily distinguish the two images in the shared layer than STL by expanding the distance between the ambiguous images.

\subsection{CIFAR-10 and CIFAR-100 Tiny Images Dataset}\label{cifar}
The CIFAR-10 dataset \cite{krizhevsky2009learning} is composed of 10 classes of natural images with 50,000 training images, and 10,000 testing images. Each image is a RGB image of size 32$\times$32. For this dataset, we applied the same CNN architecture and decoders as those in MNIST dataset, while we evaluated our method and all of the baseline methods using two settings. One setting was training the model on the original dataset without any data augmentation method. The other setting was training on the dataset with some data augmentation techniques, i.e., randomly performing horizontal flips and gray scale. Because the labels of the auxiliary tasks were generated by the whole image, we did not apply the randomly crop technique on the images.

\begin{table}[t]
\begin{center}
\scalebox{0.8}{
\begin{tabular}{l|c|c}  
\hline
\multirow{2}{*}{\textbf{Model}} & \multicolumn{2}{c}{\textbf{Acc.}} \\  
\cline{2-3}  
 & CNN & CNN$\ddagger$ \\  
\hline
STL \cite{yang2016deep} & $72.48$ & $75.91$ \\
ACAI embedding finetune \cite{berthelot2018understanding} & $64.94$ &$64.94$ \\
MTL + Tasks with random labels, $T = 4$& $70.81$ & $74.48$  \\
MTL + Tasks with $k$-means labels $T = 4$& $71.06$ & $74.61$ \\
PTA-F, $T = 4$ \cite{meyerson2018pseudo} & $69.20$ & $72.54$ \\
PTA-F, $T = 12$ \cite{meyerson2018pseudo} & $68.48$ & $70.96$ \\
\hline
\textbf{Meta-MTL}, $T = 4$ & $\mathbf{75.02}$ & $\mathbf{78.65}$ \\
\textbf{Meta-MTL}, $T = 12$ & $\mathbf{75.69}$ & $\mathbf{79.65}$ \\
\hline
\end{tabular}}
\end{center}
\caption{Accuracy on CIFAR-10 dataset. The models marked with $\ddagger$ apply data augmentation.}
\label{tab:cifar10}
\end{table}

\begin{table}[t]
\begin{center}
\scalebox{0.8}{
\begin{tabular}{l|c|c}  
\hline
\multirow{2}{*}{\textbf{Model}} & \multicolumn{2}{c}{\textbf{Acc.}} \\  
\cline{2-3}  
& 20 C & 100 C  \\  
\hline
STL \cite{yang2016deep} & $55.94$ & $44.19$  \\
ACAI embedding finetune \cite{berthelot2018understanding} & $44.37$ & $34.40$\\
MTL + Tasks with random labels & $51.00$ & $41.30$ \\
MTL + Tasks with $k$-means labels & $51.84$ & $42.30$ \\
PTA-F, $T = 8$ \cite{meyerson2018pseudo} & $51.67$ & $45.86$ \\
PTA-F, $T = 20$ \cite{meyerson2018pseudo} & $51.69$ & $47.43$ \\
\hline
\textbf{Meta-MTL}, $T = 8$ & $\mathbf{59.66}$ & $\mathbf{47.01}$ \\
\textbf{Meta-MTL}, $T = 20$ & $\mathbf{60.39}$ & $\mathbf{47.94}$ \\
\hline
\end{tabular}}
\end{center}
\caption{Accuracy on CIFAR-100 dataset. The ``20 C'' means that the 100 classes in the CIFAR-100 are grouped into 20 superclasses, and the ``100 C'' means the original 100 classes.}
\label{tab:cifar100}
\end{table}

The results are listed in Table \ref{tab:cifar10}. We can see that our method outperforms the other baseline methods by a large margin. In contrast to the PTA-F trained on the Omniglot and MNIST, when the number of decoders increased, the performance of this model dropped greatly by more than 4\%. The performance of MTL models with random labels and $k$-means labels also dropped slightly. In contrast, our model exceeds the STL by more than 3\%. In addition, the contribution of our model and data augmentation may be orthogonal, because whether or not the model uses data augmentation, our methods have similar improvements.

The CIFAR-100 dataset \cite{krizhevsky2009learning} is the same in size and format as the CIFAR-10 dataset, but it contains 100 classes. Thus, the number of images in each class is only one-tenth of the CIFAR-10 dataset. This dataset also provides another label version that the 100 classes in the CIFAR-100 are grouped into 20 superclasses. We did not tune the hyper-parameters and used the same setting as the CIFAR-10 dataset. The results are listed in Table \ref{tab:cifar100}. Our Meta-MTL also outperforms the other baseline models by more than 4\% in ``20 C'' and 3\% in ``100 C'', respectively.

\begin{figure}[t]
\centering
  \includegraphics[width=2.in]{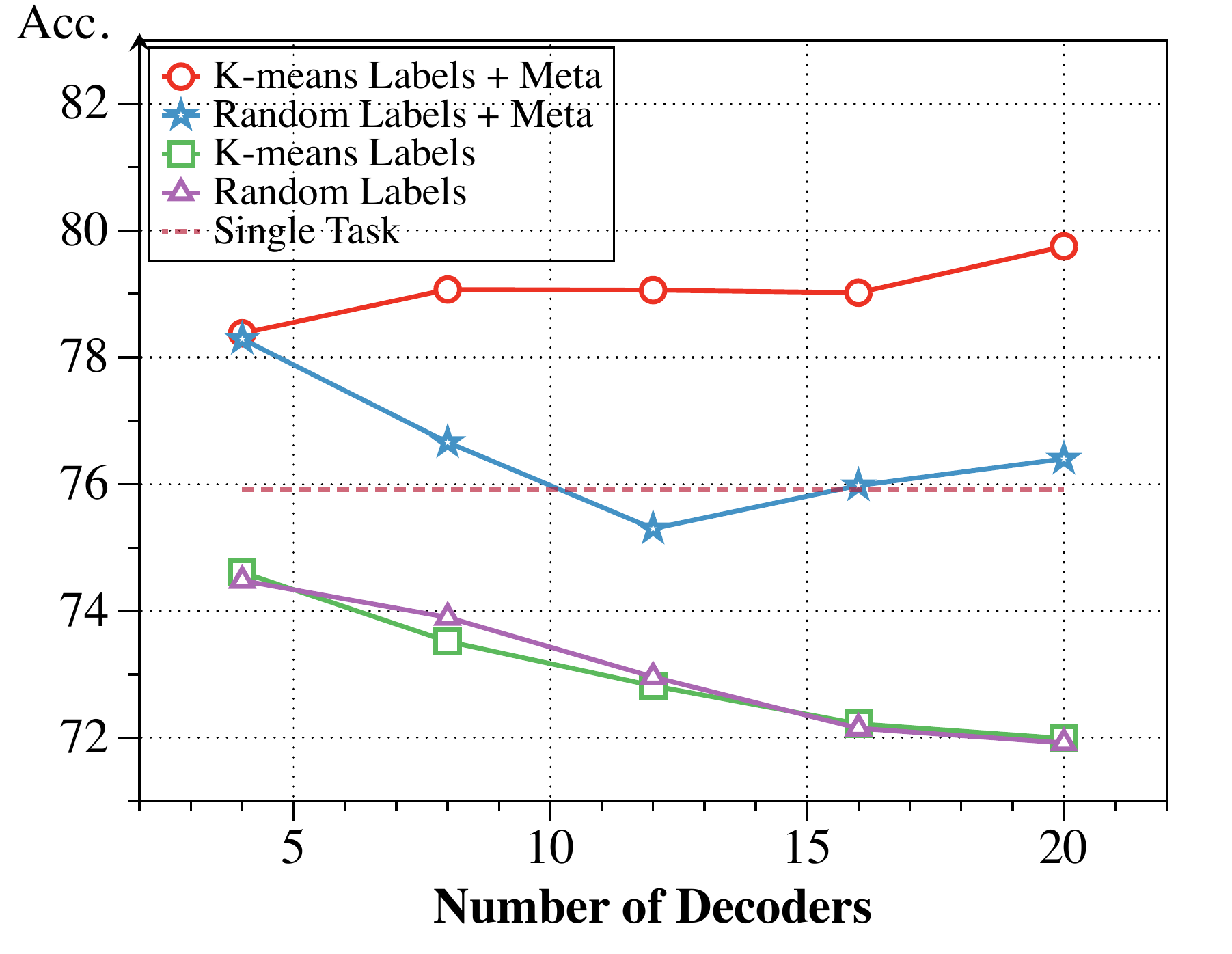}
  \caption{Comparison of MTL models trained on the CIFAR-10 datasets with different numbers of tasks.}
  \label{fig:decoders}
\end{figure}

To verify the effect of our meta-MTL algorithm and $k$-means labels, we evaluated the performance of our method and several baseline methods with different numbers of auxiliary tasks on the CIFAR-10 dataset. The comparison of these methods is shown in Figure \ref{fig:decoders}. From the figure, we can see that our model "K-means Labels + Meta" achieve a stable and outstanding performance for each number of decoders. As the number of decoders increases, the performance of the model also shows an upward trend. In contrast, the performance of the models with $k$-means or random labels but not using a meta learning algorithm gradually declines when the number of decoders increases. Comparing our method with model "K-means Lables", we can find that the proposed meta-MTL algorithm can suppress the impact of error labels and leverage ideal labels from auxiliary tasks, make the model achieve more than 8\% higher accuracy than that without the meta-learning algorithm. In addition, comparing model "K-means Labels + Meta" with "Random Labels + Meta", we can see that $k$-means labels can exhibit semantic meanings, make the model achieve more than 4\% higher accuracy.

\subsection{MiniImageNet}
In this subsection, we aim to verify the performance of the proposed method on a more difficult dataset. MiniImageNet dataset \cite{ravi2016optimization}, which consists of 100 classes, each with 600 examples of size 84$\times$84. The images are predominantly natural and realistic. We split the dataset into training and test sets in a 50,000/10,000 ratio, then trained and tested these methods using five-fold cross validation. To evaluate the performance of the proposed method based on different embedding spaces, we compared the DeepCluster \cite{caron2018deep} with ACAI \cite{berthelot2018understanding} method to produce embeddings. Moreover, we applied random scaling and randomly select half dimensions of the embeddings for the $k$-means clusters to produce different partitions. The baseline model is the same as \cite{finn2017model}. The underlying model $\mathcal{F}$ is a simple four layer convolutional network, of which each layer has 32 filters and 3$\times$3 kernels, and is followed by a 2$\times$2 maxpooling layer. Each task-specific decoder $\mathcal{D}_t$ has one fully connected layer with 100 neurons.


The final results are shown in Table \ref{fig:miniimagenet}. The Meta-MTL model still outperforms all of the baseline models. Comparing the model finetuned on the DeepCluster embeddings with that on the ACAI embeddings, we found that the DeepCluster model has a much stronger performance than the ACAI. This is because ACAI’s pixel-wise reconstruction loss may prioritize information about the background color in these complex images, while reducing the focus on the target objects \cite{hsu2018unsupervised}. Hence, the Meta-MTL model can achieve better results based on the clusters obtained by DeepCluster embeddings compared to those obtained by ACAI embeddings, which further leads to better performance of Meta-MTL. However, we could hardly distinguish which method to transform the embedding space for multiple partitions is better. The method with random scaling is not always better than that with random selecting half dimensions of embeddings. Therefore, both the random scaling and random selecting methods can produce different effective embeddings to achieve similar performances.

\begin{table}[t]
\begin{center}
\scalebox{0.8}{
\begin{tabular}{c|l|c}  
\hline
\textbf{Embedding} & \textbf{Model} & \textbf{Acc.} \\  
\hline
& STL & $38.98$ \\
\textbf{Random initialization} & MTL, Random Labels  & $36.70$ \\
 & PTA-F, $T = 4$ & $34.93$ \\
\hline
\multirow{4}{*}{\textbf{DeepCluster}} & Embedding finetune & $35.14$ \\
& MTL, $k$-means, $T = 4, \diamondsuit$ & $37.24$ \\
& MTL, $k$-means, $T = 4, \heartsuit$ & $36.78$ \\
\cline{2-3}
& \textbf{Meta-MTL}, $T = 4, \diamondsuit$ & $\mathbf{41.07}$ \\
& \textbf{Meta-MTL}, $T = 4, \heartsuit$ & $\mathbf{40.86}$ \\
\hline
\multirow{4}{*}{\textbf{ACAI}} & Embedding finetune & $22.90$ \\
& MTL, $k$-means, $T = 4, \diamondsuit$ & $37.36$ \\
& MTL, $k$-means, $T = 4, \heartsuit$ & $36.98$ \\
\cline{2-3}
& \textbf{Meta-MTL}, $T = 4, \diamondsuit$ & $\mathbf{40.43}$ \\
& \textbf{Meta-MTL}, $T = 4, \heartsuit$ & $\mathbf{40.60}$ \\
\hline
\end{tabular}}
\end{center}
\caption{Accuracy on miniImageNet dataset. The models marked with $\diamondsuit$ apply the random scaling on the embeddings to obtain the different tasks, while those marked with $\heartsuit$ apply random selection for half of the dimensions on the embeddings.}
\label{fig:miniimagenet}
\end{table}

\section{Conclusion}
In this work, we proposed the use of an unsupervised clustering method to construct a variety of auxiliary tasks to improve single task learning. We found that with simple $k$-means clustering mechanisms for partitioning the embedding space, some of the clusters could construct reasonable related auxiliary tasks. We also proposed a novel meta-learning-based multi-task learning framework to assist the fast and robust adaptation for the main task, which can effectively reduce the influence of the non-ideal clusters. Because the unsupervised clustering methods can be quickly implemented on a large scale, we can easily construct a sufficient number of auxiliary tasks. Experimental results across five image datasets demonstrated that the proposed method significantly outperforms the previous methods.

\section*{Acknowledgments}
The authors wish to thank the anonymous reviewers for their helpful comments. This work was partially funded by China National Key R\&D Program (No. 2018YFB1005104, 2018YFC0831105), National Natural Science Foundation of China (No. 61976056, 61532011, 61751201), Shanghai Municipal Science and Technology Major Project (No.2018SHZDZX01), Science and Technology Commission of Shanghai Municipality Grant  (No.18DZ1201000, 16JC1420401, 17JC1420200).

\bibliographystyle{aaai}
\bibliography{aaai}

\begin{thebibliography}{}

\bibitem[\protect\citeauthoryear{Andrychowicz \bgroup et al\mbox.\egroup
  }{2016}]{andrychowicz2016learning}
Andrychowicz, M.; Denil, M.; Gomez, S.; Hoffman, M.~W.; Pfau, D.; Schaul, T.;
  Shillingford, B.; and De~Freitas, N.
\newblock 2016.
\newblock Learning to learn by gradient descent by gradient descent.
\newblock In {\em NeurIPS},  3981--3989.

\bibitem[\protect\citeauthoryear{Baxter}{1997}]{baxter1997bayesian}
Baxter, J.
\newblock 1997.
\newblock A bayesian/information theoretic model of learning to learn via
  multiple task sampling.
\newblock {\em Machine learning} 28(1):7--39.

\bibitem[\protect\citeauthoryear{Ben-David and Borbely}{2008}]{ben2008notion}
Ben-David, S., and Borbely, R.~S.
\newblock 2008.
\newblock A notion of task relatedness yielding provable multiple-task learning
  guarantees.
\newblock {\em Machine learning} 73(3):273--287.

\bibitem[\protect\citeauthoryear{Berthelot \bgroup et al\mbox.\egroup
  }{2019}]{berthelot2018understanding}
Berthelot, D.; Raffel, C.; Roy, A.; and Goodfellow, I.
\newblock 2019.
\newblock Understanding and improving interpolation in autoencoders via an
  adversarial regularizer.
\newblock In {\em ICLR}.

\bibitem[\protect\citeauthoryear{Bingel and
  S{\o}gaard}{2017}]{bingel2017identifying}
Bingel, J., and S{\o}gaard, A.
\newblock 2017.
\newblock Identifying beneficial task relations for multi-task learning in deep
  neural networks.
\newblock In {\em EACL},  164--169.

\bibitem[\protect\citeauthoryear{Bojanowski and
  Joulin}{2017}]{bojanowski2017unsupervised}
Bojanowski, P., and Joulin, A.
\newblock 2017.
\newblock Unsupervised learning by predicting noise.
\newblock In {\em ICML},  517--526.
\newblock JMLR. org.

\bibitem[\protect\citeauthoryear{Bonilla, Chai, and
  Williams}{2008}]{bonilla2008multi}
Bonilla, E.~V.; Chai, K.~M.; and Williams, C.
\newblock 2008.
\newblock Multi-task gaussian process prediction.
\newblock In {\em Advances in neural information processing systems},
  153--160.

\bibitem[\protect\citeauthoryear{Caron \bgroup et al\mbox.\egroup
  }{2018}]{caron2018deep}
Caron, M.; Bojanowski, P.; Joulin, A.; and Douze, M.
\newblock 2018.
\newblock Deep clustering for unsupervised learning of visual features.
\newblock In {\em ECCV},  132--149.

\bibitem[\protect\citeauthoryear{Caruana}{1997}]{caruana1997multitask}
Caruana, R.
\newblock 1997.
\newblock Multitask learning.
\newblock {\em Machine learning} 28(1):41--75.

\bibitem[\protect\citeauthoryear{Chen, Weinberger, and
  Blitzer}{2011}]{chen2011co}
Chen, M.; Weinberger, K.~Q.; and Blitzer, J.
\newblock 2011.
\newblock Co-training for domain adaptation.
\newblock In {\em Advances in neural information processing systems},
  2456--2464.

\bibitem[\protect\citeauthoryear{Coates and Ng}{2012}]{coates2012learning}
Coates, A., and Ng, A.~Y.
\newblock 2012.
\newblock Learning feature representations with k-means.
\newblock In {\em Neural networks: Tricks of the trade}. Springer.
\newblock  561--580.

\bibitem[\protect\citeauthoryear{Collobert and
  Weston}{2008}]{collobert2008unified}
Collobert, R., and Weston, J.
\newblock 2008.
\newblock A unified architecture for natural language processing: Deep neural
  networks with multitask learning.
\newblock In {\em ICML},  160--167.
\newblock ACM.

\bibitem[\protect\citeauthoryear{Deng, Hinton, and
  Kingsbury}{2013}]{deng2013new}
Deng, L.; Hinton, G.; and Kingsbury, B.
\newblock 2013.
\newblock New types of deep neural network learning for speech recognition and
  related applications: An overview.
\newblock In {\em 2013 IEEE International Conference on Acoustics, Speech and
  Signal Processing},  8599--8603.
\newblock IEEE.

\bibitem[\protect\citeauthoryear{Donahue, Kr{\"a}henb{\"u}hl, and
  Darrell}{2016}]{donahue2016adversarial}
Donahue, J.; Kr{\"a}henb{\"u}hl, P.; and Darrell, T.
\newblock 2016.
\newblock Adversarial feature learning.
\newblock {\em arXiv preprint arXiv:1605.09782}.

\bibitem[\protect\citeauthoryear{Finn, Abbeel, and
  Levine}{2017}]{finn2017model}
Finn, C.; Abbeel, P.; and Levine, S.
\newblock 2017.
\newblock Model-agnostic meta-learning for fast adaptation of deep networks.
\newblock In {\em Proceedings of the 34th International Conference on Machine
  Learning-Volume 70},  1126--1135.
\newblock JMLR. org.

\bibitem[\protect\citeauthoryear{Girshick}{2015}]{girshick2015fast}
Girshick, R.
\newblock 2015.
\newblock Fast r-cnn.
\newblock In {\em Proceedings of the IEEE international conference on computer
  vision},  1440--1448.

\bibitem[\protect\citeauthoryear{Han and Zhang}{2016}]{han2016multi}
Han, L., and Zhang, Y.
\newblock 2016.
\newblock Multi-stage multi-task learning with reduced rank.
\newblock In {\em Thirtieth AAAI Conference on Artificial Intelligence}.

\bibitem[\protect\citeauthoryear{He \bgroup et al\mbox.\egroup
  }{2016}]{he2016deep}
He, K.; Zhang, X.; Ren, S.; and Sun, J.
\newblock 2016.
\newblock Deep residual learning for image recognition.
\newblock In {\em CVPR},  770--778.

\bibitem[\protect\citeauthoryear{Hsu, Levine, and
  Finn}{2019}]{hsu2018unsupervised}
Hsu, K.; Levine, S.; and Finn, C.
\newblock 2019.
\newblock Unsupervised learning via meta-learning.
\newblock In {\em International Conference on Learning Representations}.

\bibitem[\protect\citeauthoryear{Krizhevsky, Hinton, and
  others}{2009}]{krizhevsky2009learning}
Krizhevsky, A.; Hinton, G.; et~al.
\newblock 2009.
\newblock Learning multiple layers of features from tiny images.
\newblock Technical report, Citeseer.

\bibitem[\protect\citeauthoryear{Krizhevsky, Sutskever, and
  Hinton}{2012}]{krizhevsky2012imagenet}
Krizhevsky, A.; Sutskever, I.; and Hinton, G.~E.
\newblock 2012.
\newblock Imagenet classification with deep convolutional neural networks.
\newblock In {\em NeurIPS},  1097--1105.

\bibitem[\protect\citeauthoryear{Lake, Salakhutdinov, and
  Tenenbaum}{2015}]{lake2015human}
Lake, B.~M.; Salakhutdinov, R.; and Tenenbaum, J.~B.
\newblock 2015.
\newblock Human-level concept learning through probabilistic program induction.
\newblock {\em Science} 350(6266):1332--1338.

\bibitem[\protect\citeauthoryear{Li \bgroup et al\mbox.\egroup
  }{2018}]{li2018learning}
Li, D.; Yang, Y.; Song, Y.-Z.; and Hospedales, T.~M.
\newblock 2018.
\newblock Learning to generalize: Meta-learning for domain generalization.
\newblock In {\em Thirty-Second AAAI Conference on Artificial Intelligence}.

\bibitem[\protect\citeauthoryear{Long \bgroup et al\mbox.\egroup
  }{2017}]{long2017learning}
Long, M.; Cao, Z.; Wang, J.; and Philip, S.~Y.
\newblock 2017.
\newblock Learning multiple tasks with multilinear relationship networks.
\newblock In {\em Advances in neural information processing systems},
  1594--1603.

\bibitem[\protect\citeauthoryear{Luong \bgroup et al\mbox.\egroup
  }{2015}]{luong2015multi}
Luong, M.-T.; Le, Q.~V.; Sutskever, I.; Vinyals, O.; and Kaiser, L.
\newblock 2015.
\newblock Multi-task sequence to sequence learning.
\newblock {\em arXiv preprint arXiv:1511.06114}.

\bibitem[\protect\citeauthoryear{Meyerson and
  Miikkulainen}{2018}]{meyerson2018pseudo}
Meyerson, E., and Miikkulainen, R.
\newblock 2018.
\newblock Pseudo-task augmentation: From deep multitask learning to intratask
  sharing-and back.
\newblock In {\em International Conference on Machine Learning},  3508--3517.

\bibitem[\protect\citeauthoryear{Ravi and
  Larochelle}{2016}]{ravi2016optimization}
Ravi, S., and Larochelle, H.
\newblock 2016.
\newblock Optimization as a model for few-shot learning.

\bibitem[\protect\citeauthoryear{Rosenberg, Hebert, and
  Schneiderman}{2005}]{rosenberg2005semi}
Rosenberg, C.; Hebert, M.; and Schneiderman, H.
\newblock 2005.
\newblock Semi-supervised self-training of object detection models.
\newblock {\em WACV/MOTION} 2.

\bibitem[\protect\citeauthoryear{Ruder}{2017}]{ruder2017overview}
Ruder, S.
\newblock 2017.
\newblock An overview of multi-task learning in deep neural networks.
\newblock {\em arXiv preprint arXiv:1706.05098}.

\bibitem[\protect\citeauthoryear{Simonyan and
  Zisserman}{2014}]{simonyan2014very}
Simonyan, K., and Zisserman, A.
\newblock 2014.
\newblock Very deep convolutional networks for large-scale image recognition.
\newblock {\em arXiv preprint arXiv:1409.1556}.

\bibitem[\protect\citeauthoryear{Snell, Swersky, and
  Zemel}{2017}]{snell2017prototypical}
Snell, J.; Swersky, K.; and Zemel, R.
\newblock 2017.
\newblock Prototypical networks for few-shot learning.
\newblock In {\em Advances in Neural Information Processing Systems},
  4077--4087.

\bibitem[\protect\citeauthoryear{Vinyals \bgroup et al\mbox.\egroup
  }{2016}]{vinyals2016matching}
Vinyals, O.; Blundell, C.; Lillicrap, T.; Wierstra, D.; et~al.
\newblock 2016.
\newblock Matching networks for one shot learning.
\newblock In {\em Advances in neural information processing systems},
  3630--3638.

\bibitem[\protect\citeauthoryear{Wilson \bgroup et al\mbox.\egroup
  }{2007}]{wilson2007multi}
Wilson, A.; Fern, A.; Ray, S.; and Tadepalli, P.
\newblock 2007.
\newblock Multi-task reinforcement learning: a hierarchical bayesian approach.
\newblock In {\em ICML},  1015--1022.
\newblock ACM.

\bibitem[\protect\citeauthoryear{Yang and Hospedales}{2016}]{yang2016deep}
Yang, Y., and Hospedales, T.
\newblock 2016.
\newblock Deep multi-task representation learning: A tensor factorisation
  approach.
\newblock {\em arXiv preprint arXiv:1605.06391}.

\bibitem[\protect\citeauthoryear{Yim \bgroup et al\mbox.\egroup
  }{2015}]{yim2015rotating}
Yim, J.; Jung, H.; Yoo, B.; Choi, C.; Park, D.; and Kim, J.
\newblock 2015.
\newblock Rotating your face using multi-task deep neural network.
\newblock In {\em Proceedings of the IEEE Conference on Computer Vision and
  Pattern Recognition},  676--684.

\bibitem[\protect\citeauthoryear{Zhang and Yang}{2017}]{zhang2017survey}
Zhang, Y., and Yang, Q.
\newblock 2017.
\newblock A survey on multi-task learning.
\newblock {\em arXiv preprint arXiv:1707.08114}.

\end{thebibliography}

\end{document}